\documentclass[letterpaper, 10 pt, conference]{ieeeconf}

\IEEEoverridecommandlockouts
\usepackage{amsmath,amssymb,amsfonts}
\usepackage{booktabs}
\usepackage{graphicx}
\usepackage{multirow}
\usepackage{placeins}
\usepackage{flafter}
\let\labelindent\relax
\usepackage{enumitem}
\usepackage[table]{xcolor}
\usepackage{colortbl}
\usepackage[hidelinks]{hyperref}
\graphicspath{{figure/}}

\title{\LARGE \bf Beyond State-as-Action: Exploiting Command--State Discrepancy for Robot Imitation Learning}

\author{Peiyan Li$^{1,6,\dagger}$, Yueran Tao$^{1,6,\dagger}$, Enhao Zhang$^{2,6,\dagger}$, Zhixuan Zhao$^{1,6}$,\\
Chenghao Yue$^{1,6}$, Hao Wang$^{3,6}$, Lei Lv$^{4,6}$, Wentao Zhao$^{1,6}$, Jiahao Chen$^{5,6}$,\\
Xin Liu$^{1,6}$, Kangyao Huang$^{1,6}$, Yu Luo$^{1,6,*}$, Huaping Liu$^{1,6,*}$\thanks{$^{1}$Tsinghua University; $^{2}$Imperial College London; $^{3}$Dalian University of Technology; $^{4}$Tongji University; $^{5}$Peking University; $^{6}$SEEN\textperiodcentered E Robotics.}\thanks{$^{\dagger}$Equal contribution (co-first authors).}\thanks{$^{*}$Co-corresponding authors: Yu Luo and Huaping Liu.}}

\makeatletter
\def\@maketitle{\newpage
  \begin{center}
    \vspace*{0pt}
    {\LARGE\bfseries\@title\par}
    \vspace{6pt}
    {\fontsize{10}{12}\selectfont\@author\par}
  \end{center}}
\def\@IEEEdynamictitlevspace{\vspace{3pt}}
\makeatother

\begin{document}
\maketitle
\thispagestyle{empty}
\pagestyle{empty}
\bstctlcite{icra:BSTcontrol}

\begin{abstract}
Constructing action targets from measured robot motion is an established approach in imitation learning. Under interaction constraints, however, command--state discrepancy may reflect control demands that motion alone does not capture. We investigate when this information matters and how to exploit it. Across three real-robot tasks, task and phase analyses reveal larger supervision gaps under constrained interaction, while selective command retention provides evidence of locally useful command information. Building on these findings, we propose Command--State Discrepancy Weighting (CSDW), which accounts for robot response times and combines subsequent progress, persistent unmet demand, and demand changes into continuous weights for command supervision. The method requires no task-phase annotations or changes to policy architecture or inference. CSDW improves over uniform command supervision on constrained tasks, while methods perform similarly in the less constrained task. Project page: \url{https://seen-e.github.io/CSDW/}.
\end{abstract}

\section{INTRODUCTION}

Measured robot motion provides a practical basis for action
supervision in imitation learning. Existing training pipelines
use measured states or their changes as action
targets~\cite{walke2023bridgedata,kim2026race}, a choice we refer
to as \emph{state-as-action} supervision. These targets describe
the motion achieved during demonstration, allowing a tracking
controller to reproduce it during execution. When paired with
a suitable controller, this representation can reduce tracking
errors under changes in execution timing~\cite{kim2026race}.

\begin{figure}[!t]
\centering
\includegraphics[width=\columnwidth]{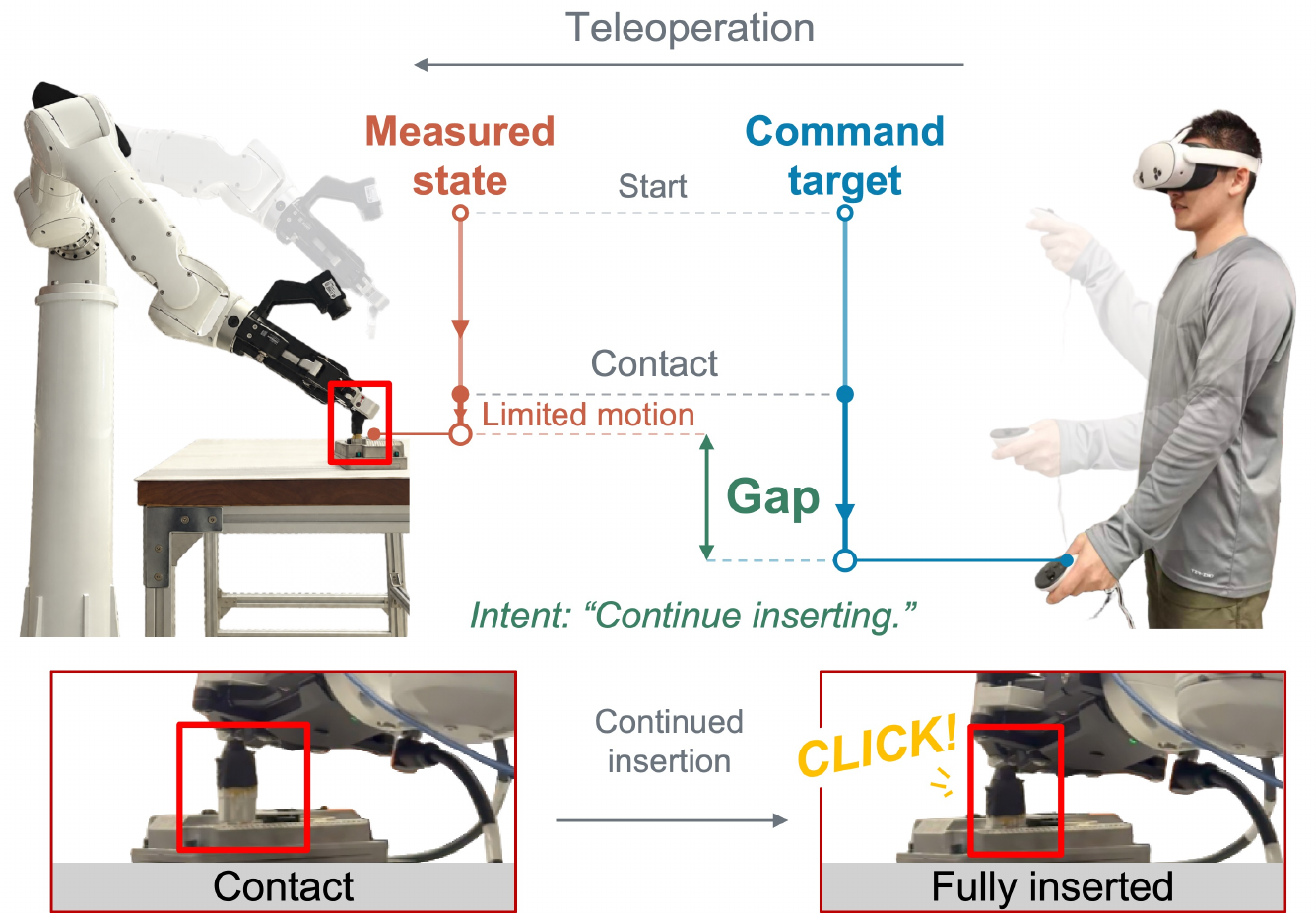}
\caption{Control intent in command--state discrepancy.
During plug insertion, contact limits further robot motion
despite continued insertion commands. The gap between the command target and the robot's current measured state can indicate the operator's intent to continue insertion, even when actual motion is limited.
Faded poses indicate earlier configurations, and the lower
images show contact followed by full insertion.}
\label{fig:motivation}
\vspace{-1 em}
\end{figure}

The usefulness of retaining the original commands can depend
on the interaction. During plug insertion, for example, contact
limits further robot motion while the operator continues
commanding insertion (Fig.~\ref{fig:motivation}).
Under impedance control, a maintained offset between the
commanded and actual position can sustain contact force
despite limited displacement. Such a discrepancy can be both
a tracking error and a cue to the operator's \emph{control intent}
to continue inserting~\cite{chen2025dexforce,surendran2026bridge}.
The usefulness of this information may vary across tasks and
across transport, alignment, and insertion within a demonstration.
We ask when it benefits imitation learning and how the temporal
relationship between commands and robot responses can guide
training.

We study these questions using paired recordings from the same
demonstrations, comparing recorded joint commands with future
measured joint positions under a shared control interface.
We evaluate on three real-robot tasks with varying interaction
constraints: cable insertion, connector mating, and package
transfer. Overall success and intermediate outcomes allow us
to examine how the usefulness of command information varies
across tasks and interaction phases.
To identify where command information is useful, we use \emph{Hybrid} supervision. It retains recorded arm commands at frames selected by a temporal discrepancy score and uses state-as-action targets elsewhere.

These analyses motivate giving additional training emphasis
to moments when commands may carry useful control information.
The aim is to help the policy learn when to sustain a request
despite limited motion and when to adjust or relax it.
We therefore propose \emph{Command--State Discrepancy Weighting}
(CSDW), which assigns continuous training weights by considering
how commands are maintained or revised and how the robot
subsequently responds. Estimated response times set the
timescale for examining these patterns and their changes
throughout a demonstration. CSDW retains the recorded command
targets and computes the weights offline, without task-phase
annotations or changes to policy architecture or inference.

Across the three tasks, Command outperforms State-as-Action most clearly in the two insertion tasks, whereas the two perform similarly on package transfer. The gap is largest at final completion, where sustained interaction constraints are most pronounced. Hybrid preserves much of this benefit while retaining commands only at discrepancy-selected frames, providing evidence of locally useful command information. CSDW further improves over uniform Command on both insertion tasks, with only a modest gain on package transfer. Together, these results show that the value of command-state discrepancy depends on both the interaction condition and the task phase.

Our contributions are:
\begin{itemize}[leftmargin=*, labelindent=0pt, label=\textbullet]

\item \textbf{Task dependence of command supervision.}
Across three real-robot tasks, command supervision provides
larger completion gains over state-as-action supervision
in tasks involving sustained interaction constraints,
with similar final success on package transfer.

\item \textbf{Command information within demonstrations.}
Stage outcomes show larger supervision gaps at completion
than at alignment. Retaining arm commands only in
discrepancy-selected segments preserves the observed
benefit of full command supervision.

\item \textbf{Training with command--state discrepancy.}
We propose CSDW, an offline method that weights the original
command supervision loss using temporal discrepancy
evidence, improving completion over uniform command
supervision without changing policy architecture or inference.

\end{itemize}

\section{Related Work}
\label{sec:related-work}

\subsection{Action Supervision in Robot Learning}

Imitation learning trains policies to reproduce actions in demonstration datasets~\cite{chi2024diffusionpolicy,black2025pi05}. What these actions represent depends on data collection and preprocessing. Early bilateral-control imitation learning distinguished command and response signals, using command-side prediction to preserve control information not fully reflected in measured responses~\cite{sasagawa2020bilateral}. DROID records commands separately from measured states~\cite{khazatsky2024droid}, while ALOHA and GELLO use leader joint positions to command follower robots~\cite{zhao2023act,wu2024gello}. Open X-Embodiment combines datasets with heterogeneous action interfaces~\cite{openx2024}, making the source of action labels relevant to their interpretation.

Other pipelines construct targets from demonstrated motion. The released BridgeData V2 training pipeline relabels arm actions using changes in measured end-effector pose~\cite{walke2023bridgedata}. In simulation, ACT extracts arm joint trajectories from scripted rollouts and replays them as position commands, while retaining gripper commands~\cite{zhao2023act}. RACE uses reached states as targets together with controller and timing adjustments for faster execution~\cite{kim2026race}. UMI learns relative end-effector trajectories from handheld demonstrations and addresses execution latency~\cite{chi2024umi}. State-only imitation instead recovers missing expert actions through learned inverse dynamics~\cite{radosavovic2021soil}, rather than directly using measured states as action targets.

\subsection{Control Information in Demonstrations}

Physical constraints can make observed motion an incomplete description of control demand. DexForce constructs force-informed position targets from kinesthetic demonstrations~\cite{chen2025dexforce}, while Adaptive Compliance Policy learns virtual targets and stiffness from force--motion data~\cite{hou2025acp}. RealDexUMI reports degraded performance when recorded hand commands are replaced by measured hand states~\cite{xu2026realdexumi}. BRIDGE combines handheld and teleoperated supervision through state-gated experts~\cite{surendran2026bridge}, addressing differences in their suitability across interaction phases.

Complementary approaches enrich sensing or control. ForceMimic and UMI-FT learn force or compliance targets from force--motion demonstrations~\cite{liu2025forcemimic,choi2026umift}, while ForceVLA2 predicts hybrid force--position actions~\cite{li2026forcevla2}. Reactive Diffusion Policy and ViTaL incorporate tactile feedback~\cite{xue2025rdp,zhao2025vital}, while FILIC combines estimated wrench observations with impedance control~\cite{ge2026filic}. We examine how the choice between command and state supervision affects learning across tasks and task phases, while keeping the control interface fixed.

\subsection{Data Weighting for Imitation Learning}

Weighting and curation methods allocate supervision according to different signals. Learned weighting uses discriminator and policy estimates~\cite{wang2021weight}, or stationary-distribution correction and weighted behavior cloning~\cite{kim2022demodice}, to exploit imperfect demonstrations. DemInf scores trajectories using state--action mutual information~\cite{hejna2025deminf}, while CUPID uses evaluation rollouts to estimate their influence on policy return~\cite{agia2025cupid}. IWR upweights human interventions~\cite{mandlekar2020iwr}; SIRIUS also reduces the contribution of pre-intervention robot actions~\cite{liu2023sirius}. Our approach weights command supervision using the temporal relationship between recorded commands and measured responses.

\begin{figure*}[t]
  \centering
  \includegraphics[width=\textwidth]{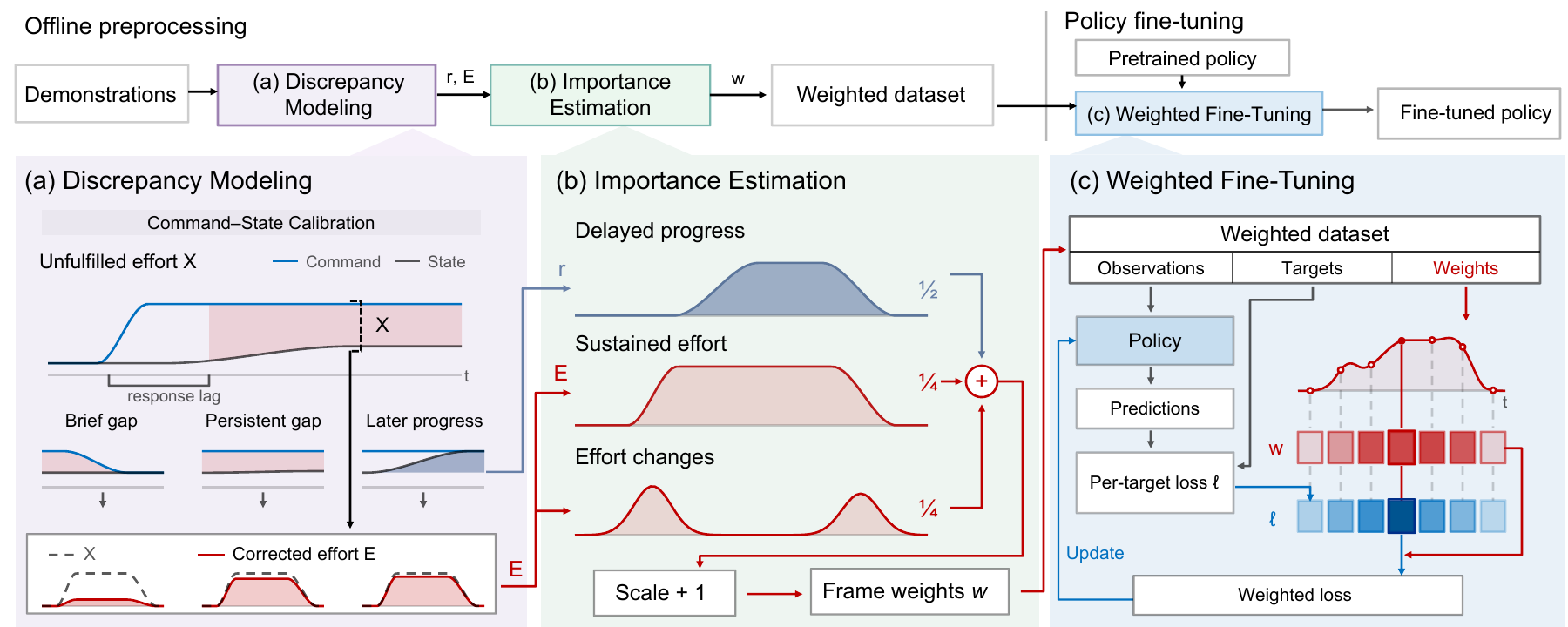}
  \caption{CSDW overview. (a) Paired command--state trajectories provide
unfulfilled effort $X$, corrected effort $E$, and separate response cues $r$.
(b) Delayed progress $C_r$ is obtained from $r$; sustained effort $C_l$ and
effort changes $C_e$ are obtained from $E$. The three profiles determine
frame weights $w$. (c) Weighted fine-tuning applies each target frame's
cached weight to its original policy loss, retaining command targets.
Scoring is offline; policy inputs and inference are unchanged.
Curves are schematic.}
  \label{fig:csdw-framework}
\end{figure*}

\section{Problem Formulation}
\label{sec:problem-formulation}

We consider demonstrations containing synchronized policy observations
$x_t$, recorded controller commands, and measured robot states. The
observations include images, proprioception, and the task instruction.
Command supervision retains the recorded commands as targets $a_t^C$
in the policy's action representation, whereas \emph{State-as-Action}
constructs targets from measured motion. These targets describe the
requested behavior and the observed response, respectively.

For discrepancy analysis, $u_t$ and $s_t$ denote paired command and
state signals calibrated to account for channel ranges and state polarity. This
calibration is used for scoring; the original command targets $a_t^C$
are retained for training. For each paired channel $j$, their signed
difference defines the command--state discrepancy,
\begin{equation}
e_{t,j}=u_{t,j}-s_{t,j}.
\label{eq:command-state-discrepancy}
\end{equation}
For a future-frame offset $k\geq0$ within the same demonstration, the
command change and state response are
\begin{equation}
\begin{aligned}
\Delta u_{t,j}(k)&=u_{t+k,j}-u_{t,j},\\
\Delta s_{t,j}(k)&=s_{t+k,j}-s_{t,j}.
\end{aligned}
\label{eq:command-state-changes}
\end{equation}
The later discrepancy is therefore
\begin{equation}
e_{t+k,j}=e_{t,j}+\Delta u_{t,j}(k)-\Delta s_{t,j}(k).
\label{eq:discrepancy-response}
\end{equation}
When a command is held, $\Delta u_{t,j}(k)=0$, so the discrepancy
changes only through the state response. When the command changes,
a smaller discrepancy magnitude can reflect either state progress or
a reduced request. A maintained command can also leave a persistent
residual when motion is limited. These cases motivate examining command
persistence and state response together when assigning training weights.

\section{Command--State Discrepancy Weighting}
\label{sec:methods}

CSDW computes frame weights from demonstrations containing paired
controller commands and measured robot states.
As shown in Fig.~\ref{fig:csdw-framework}, we first examine command
persistence and later state responses to construct unfulfilled and
corrected effort scores, together with separate response cues.
Delayed progress, sustained effort, and effort changes then determine
the additional training emphasis at each frame.
The weights are computed offline and remain fixed during fine-tuning,
where they scale the original per-target policy loss while retaining
recorded commands as training targets $a_t^C$.

\begin{figure}[t]
  \centering
  \includegraphics[width=\columnwidth]{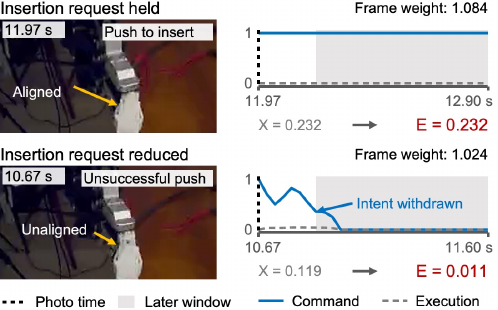}
\caption{Temporal evidence for held and reduced insertion requests.
The blue solid and gray dashed curves show retained command $c$ and directional state progress $\rho$, respectively, normalized by the initial command--state gap and
clipped to $[0,1]$. Zero $\rho$ does not necessarily imply a stationary robot. $X$ and $E$ are channel-wise unfulfilled and
corrected effort scores, not measured forces; $w$ is the final frame weight after pooling and temporal processing.
Images provide context only.}
  \label{fig:csdw-request-support}
\end{figure}

\begin{figure*}[t]
  \centering
  \includegraphics[width=\textwidth]{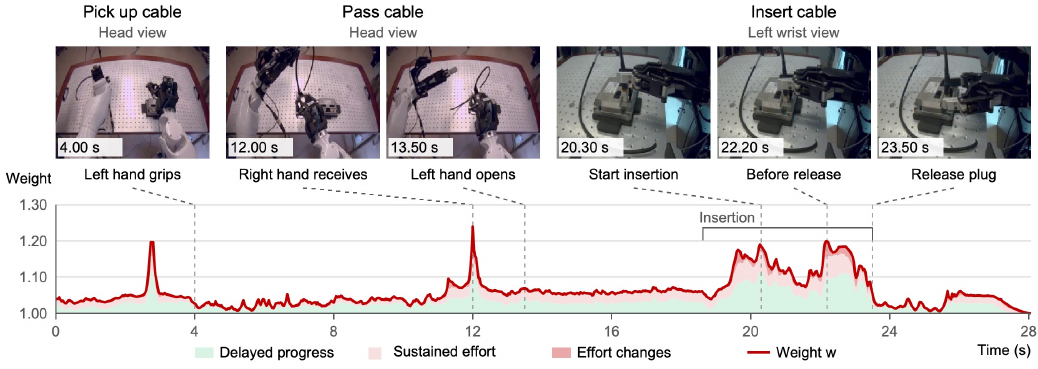}
  \caption{CSDW weights in a cable-insertion demonstration. Robot views
  contextualize pickup, handover, and insertion. Colored areas show the contributions of
  delayed progress, sustained effort, and effort changes above the unit
  baseline. Red shows the final weight.
  Additional weight extends across the annotated insertion interval,
  with local variation within it. Phase labels are manual annotations
  used only to interpret this qualitative example.}
  \label{fig:csdw-cable-weights}
\end{figure*}

\subsection{Discrepancy Modeling}
\label{subsec:discrepancy-modeling}

An insertion request may remain active despite limited state progress
or be subsequently reduced (Fig.~\ref{fig:csdw-request-support}).
We evaluate a command--state gap using command persistence and later
behavior to construct discrepancy-derived \emph{effort} scores.

\emph{Unfulfilled effort.}
Training trajectories provide the channel ranges, state polarity, and
estimated response times used for calibration and temporal analysis.
The calibrated command $u_t$ and state $s_t$ give the signed gap
$e_{t,j}=u_{t,j}-s_{t,j}$ at frame $t$ in paired channel $j$.
The estimated response times set the analysis windows; the state stream
is not shifted before computing $e_{t,j}$.

For a future-frame offset $k$, $\Delta u_{t,j}(k)$ and
$\Delta s_{t,j}(k)$ denote changes in command and state from frame $t$
to $t+k$ within the same demonstration. For an initial gap whose magnitude
exceeds the channel's noise floor, the clipped ratios
$\rho(k)=[\Delta s_{t,j}(k)/e_{t,j}]_0^1$ and
$c(k)=[1+\Delta u_{t,j}(k)/e_{t,j}]_0^1$ describe progress along the
initial gap direction and the request retained by a later command.
Here $[\cdot]_0^1$ clips to $[0,1]$; the $(t,j)$ indices of $\rho$ and
$c$ are suppressed. Both quantities are fractions of the initial gap.

The \emph{unfulfilled effort} $X_{t,j}$ multiplies four factors: a
gap-magnitude score computed from $|e_{t,j}|$ above the noise floor,
command persistence measured by the mean of $c(k)$ over the response
interval, the unresolved fraction
$1-\rho(k)$ at its end, and a discount for rapid state response. The
unresolved fraction measures how much of the initial
request has not been realized by state motion; the
rate discount compares directional progress per frame with the channel's
observed response rates. Together, they reduce emphasis on prompt
tracking while retaining evidence of continuing, unresolved requests.
Gap magnitudes at or below the noise floor contribute no unfulfilled effort.

\emph{Corrected effort.}
We use the same $c(k)$ and $\rho(k)$ over a later window to assess
support for the initial request. Response evidence $p_{t,j}^{\mathrm{st}}$
is the geometric mean of the peak of $c(k)\rho(k)$ and the
command-weighted mean of $\rho(k)$. The later gap $e_{t+k,j}$
provides complementary residual evidence:
$h_{t,j}$ averages the clipped ratio $[e_{t+k,j}/e_{t,j}]_0^1$ over
the later window, measuring how much of the initial gap remains.
The \emph{corrected effort} is
\begin{equation}
E_{t,j}=X_{t,j}\left[1-(1-p_{t,j}^{\mathrm{st}})(1-h_{t,j})\right].
\label{eq:csdw-corrected-effort}
\end{equation}
Later progress supports a request that eventually receives a response,
while a persistent residual supports one that remains unresolved.
In Fig.~\ref{fig:csdw-request-support}, state progress along the initial
gap direction is small in both examples. The held command preserves
residual support $h$, giving $E\approx X$; reducing the command weakens
this support and attenuates $E$. This correction concerns the earlier
request: a subsequent release command remains a training target, and
changes in $E$ contribute separately to weighting.

Pooling $E_{t,j}$ across channels after this correction yields the
frame-level corrected effort $E_t$. In parallel, we combine the initial evidence
with peak and averaged later progress to construct frame-level response
cues, denoted collectively by $r_t$. These cues are computed separately
from $E_t$.

\subsection{Importance Estimation}
\label{subsec:importance-estimation}

We convert the corrected effort $E_t$ and response cues $r_t$ into
three complementary temporal profiles.
\emph{Delayed progress} $C_r$ is formed from $r_t$ and captures later
state progress associated with the initial request.
\emph{Sustained effort} $C_l$ is formed from $E_t$ and retains the level of
corrected effort, whether supported by later progress or a persistent
residual. \emph{Effort changes} $C_e$ captures increases and decreases in
a smoothed $E_t$, weighted by its current or preceding level so that small
fluctuations do not receive the same emphasis as substantial changes.

We calibrate the cue scales and spread their values to neighboring frames
over the typical estimated response time.
We combine the profiles into frame weights as
\begin{equation}
w_t=1+\gamma\left[\tfrac12 C_r(t)+\tfrac14 C_l(t)+\tfrac14 C_e(t)\right]_0^1,
\label{eq:csdw-frame-weight}
\end{equation}
where $\gamma\geq0$ controls additional emphasis.
Additive fusion allows sustained effort and effort changes to
contribute even when later state progress is limited.
The coefficients are fixed design choices, not equal realized
contributions. In the cable-insertion example, additional weight extends
across the annotated insertion interval and varies within it
(Fig.~\ref{fig:csdw-cable-weights}).
The score uses no task-phase annotations, image semantics, or success
labels; its learning value is tested experimentally.

\begin{figure*}[t]
\centering
\includegraphics[width=\textwidth]{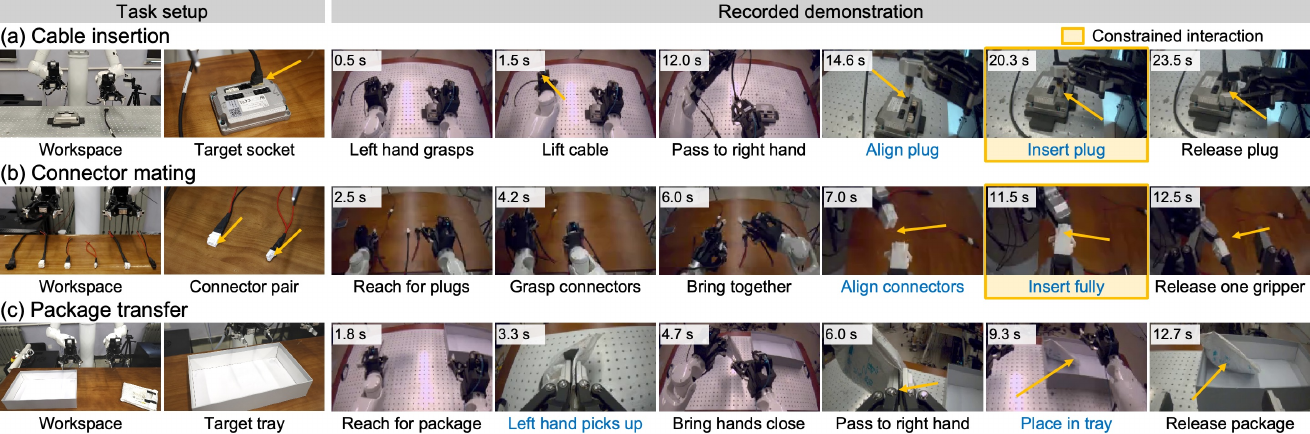}
\caption{Task setups and representative demonstrations for
(a) cable insertion, (b) connector mating, and (c) package transfer.
The first two columns show the workspace and task objects; the remaining
six show successive task stages using head and wrist views.
Blue labels indicate the evaluated stages: alignment and full insertion for (a) and (b), and pickup and placement for (c).
Gold shading marks selected insertion stages involving sustained
constrained interaction. Arrows highlight relevant objects and
interaction points. Timestamps are relative to each demonstration.}
\label{fig:experiment-tasks}
\end{figure*}

\begin{figure}[t]
\centering
\includegraphics[width=\columnwidth]{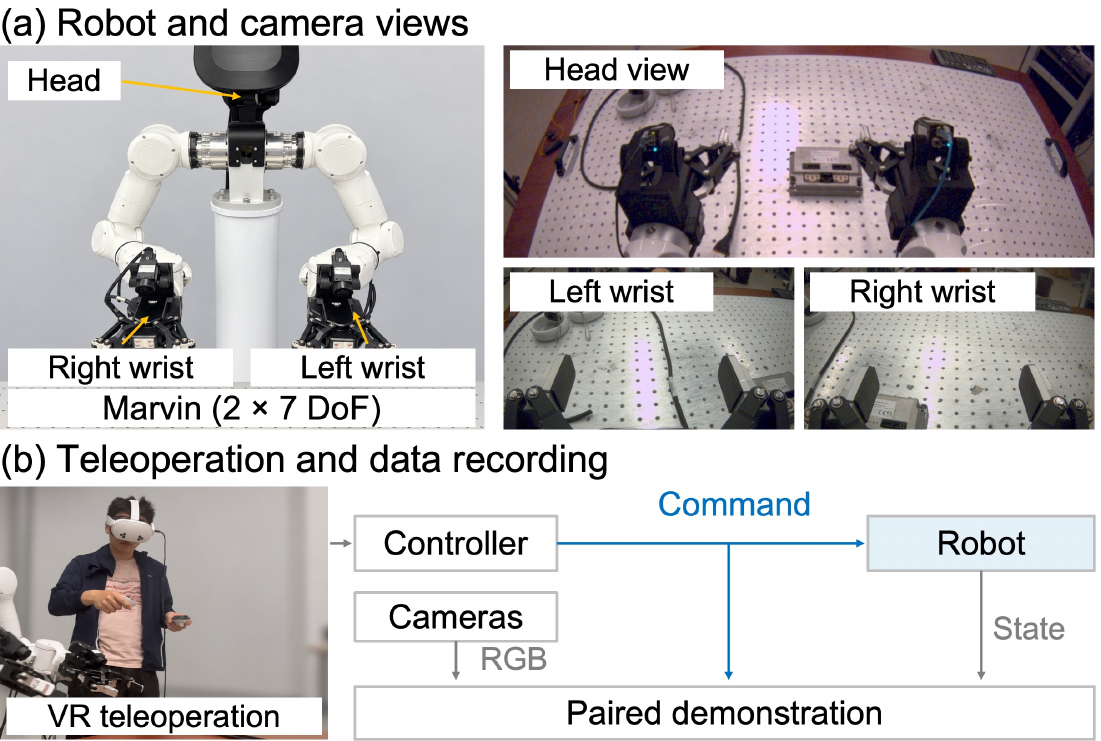}
\caption{Robot setup and demonstration collection.
(a) The Marvin robot, camera locations, and corresponding RGB views;
left and right refer to the robot.
(b) VR teleoperation and the recording of aligned commands, measured
states, and RGB images.}
\label{fig:hardware-collection}
\vspace{-1 em}
\end{figure}

\subsection{Weighted Fine-Tuning}
\label{subsec:weighted-fine-tuning}
\label{subsec:training-objective}

The weighted dataset retains the original observations $x_t$ and command
targets $a_t^C$, together with the frame weights $w_t$. With policy
parameters $\theta$, let $\ell_{b,n}(\theta)$ be the original per-target
policy loss for target $n$ of training example $b$, averaged over valid
action coordinates.
For $B$ examples and $H$ targets per example, we optimize
\begin{equation}
\mathcal L(\theta)=\frac{1}{BH}\sum_{b=1}^{B}\sum_{n=0}^{H-1}
w_{i_b,q_{b,n}}\,\ell_{b,n}(\theta),
\label{eq:csdw-training-loss}
\end{equation}
where $i_b$ identifies the demonstration and $q_{b,n}$ the frame of that
supervised target. Thus a target retains its frame weight across overlapping
action chunks. Weights are fixed and receive no gradient.
Since $w\geq1$, baseline supervision remains present.
The loss uses the ordinary mean, without normalizing weights to unit mean
or dividing by their sum; it changes both supervision allocation and
overall loss scale. Our implementation weights the original flow-matching
loss. Future states are accessed only for offline scoring, leaving the
policy architecture, observations, and inference unchanged.

\section{Experiments}
\label{sec:experiments}

Using the three tasks illustrated in Fig.~\ref{fig:experiment-tasks}, we examine when recorded commands provide useful supervision beyond
measured motion and how temporal discrepancy can guide learning from this
information. Our experiments test three hypotheses:

\begin{itemize}[leftmargin=*, labelindent=0pt, label=\textbullet]

\item \textbf{H1: Task and phase dependence of action supervision.}
The benefit of command supervision over state supervision
varies across tasks and interaction phases, with a larger
advantage during sustained constrained interaction.

\item \textbf{H2: Local value of command information.}
Retaining command targets in selected trajectory segments
can preserve the main performance benefits of full command
supervision while using state targets elsewhere.

\item \textbf{H3: Temporal discrepancy weighting.}
The temporal structure of command--state discrepancy can
guide training weights that improve policy performance
over uniform command supervision.

\end{itemize}

\subsection{Experimental Setup}
\label{subsec:experimental-setup}

\emph{Platform and demonstrations.}
We use the Marvin dual-arm robot with a head camera and a camera on each
wrist. Fig.~\ref{fig:hardware-collection} shows the platform, camera views,
and collection pipeline. An operator controls the robot through VR
teleoperation; recorded commands, measured states, and RGB images are
aligned to form paired demonstrations. The training data comprise 482
demonstrations for cable insertion, 1,044 for connector mating, and 151
for package transfer. For each task, all supervision settings are constructed
from the same paired recordings. All settings use
$\pi_{0.5}$~\cite{black2025pi05} with a flow matching objective and
RGB images, proprioception, and a task instruction as policy inputs.
CSDW weights are computed offline; the policy architecture and inputs at
deployment remain unchanged.

\begin{table*}[t]
\vspace*{1 em} %
\centering
\small
\caption{Stage success rates across three tasks.}
\label{tab:main-results}
\setlength{\tabcolsep}{4pt}
\renewcommand{\arraystretch}{1.10}
\begin{tabular*}{\textwidth}{@{\extracolsep{\fill}}
lc>{\columncolor{gray!15}}cc>{\columncolor{gray!15}}ccc@{}}
\hline
& \multicolumn{2}{c}{Cable insertion (35)}
& \multicolumn{2}{c}{Connector mating (30)}
& \multicolumn{2}{c}{Package transfer (30)} \\
\cline{2-3}\cline{4-5}\cline{6-7}
\noalign{\vskip\arrayrulewidth}
Method & Alignment & Full insertion
& Alignment & Full insertion & Pickup & Placement \\
\hline
State   & 77.1\% & 25.7\% & 26.7\% &  0.0\% &  96.7\% & 80.0\% \\
Command & 77.1\% & 60.0\% & 36.7\% & 33.3\% &  86.7\% & 80.0\% \\
Hybrid  & 71.4\% & 71.4\% & 36.7\% & 36.7\% &  93.3\% & 76.7\% \\
CSDW    & \textbf{94.3\%} & \textbf{94.3\%}
        & \textbf{73.3\%} & \textbf{73.3\%}
        & \textbf{100.0\%} & \textbf{83.3\%} \\
\hline
\end{tabular*}
\par\vspace{3pt}
\begin{minipage}{\textwidth}
\footnotesize\raggedright
\textit{Note.} Trial counts per method are given in parentheses;
light gray shading marks insertion stages involving
sustained constrained interaction.
\end{minipage}
\vspace{-2em}
\end{table*}

\begin{figure}[t]
    \centering
    \includegraphics[width=1\linewidth]{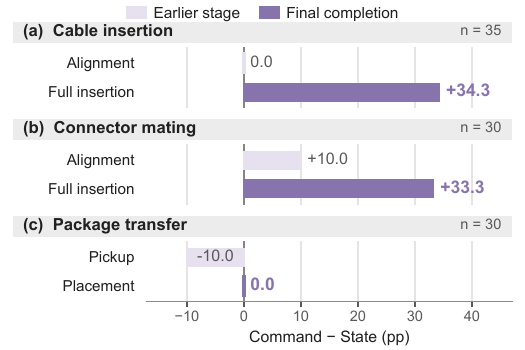}
    \caption{Command--State success-rate differences (percentage points) for (a) cable insertion, (b) connector mating, and (c) package transfer. Light purple indicates the earlier stage; dark purple indicates final completion. Positive values favor Command. All stage rates use all trials per method.}
    \label{fig:task_phrase_dependence}
\end{figure}

\begin{figure}[t]
\centering
\includegraphics[width=\columnwidth]{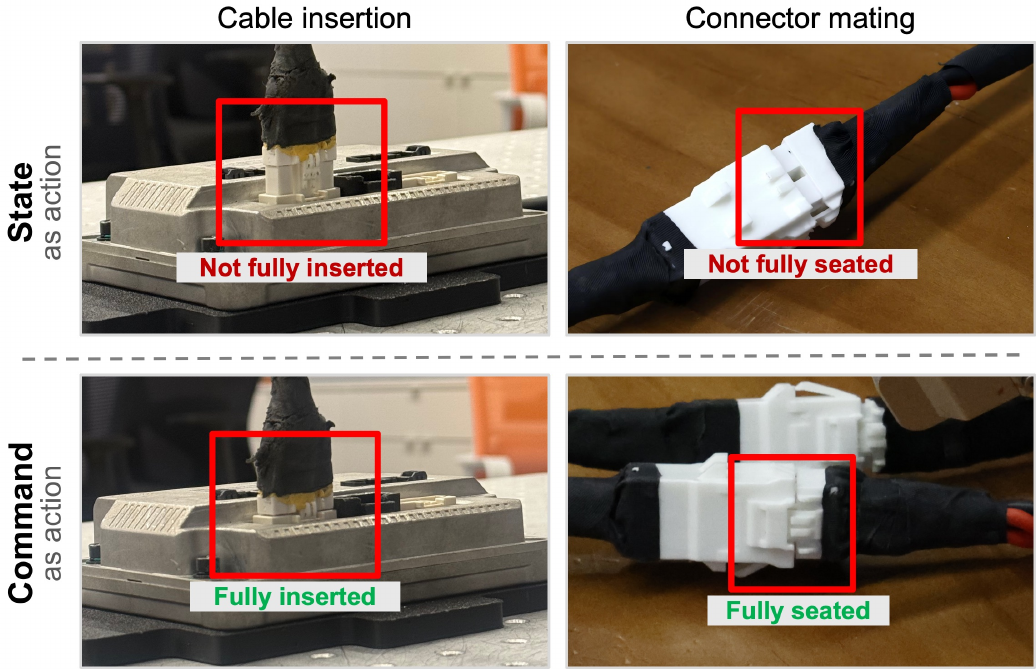}
\caption{Representative State and Command execution outcomes in the two
insertion tasks. Close-ups show partial engagement under State supervision
and full engagement under Command supervision, illustrating the distinction
between alignment and final completion in Table~\ref{tab:main-results}.}
\label{fig:experiment-behavior}
\end{figure}

\emph{Supervision settings.}
State-as-Action (State) uses future measured joint positions as arm targets,
whereas Command-as-Action (Command) uses recorded joint commands.
Hybrid retains Command targets at frames selected by CSDW scores and uses
State targets elsewhere. State, Command, and Hybrid use the ordinary policy loss.
CSDW uses the same targets as Command and scales their loss contributions
with the offline frame weights. The Command--State comparison establishes
the effect of supervision source. Hybrid is compared with State to assess
the value of selected commands and with Command to assess how much of
the full supervision benefit is retained. CSDW is compared with Command
to evaluate weighting when all command targets are already available.

\emph{Tasks and success criteria.}
Fig.~\ref{fig:experiment-tasks} illustrates the task setups
and sequences. We evaluate each supervision setting on cable
insertion, connector mating, and package transfer with 35, 30,
and 30 trials, respectively. Across the four supervision
settings, this amounts to 380 real-robot trials over
approximately six hours of evaluation sessions.

In \emph{cable insertion}, the left hand picks up a cable
and transfers it to the right hand while adjusting its pose.
The right hand aligns the plug with a fixed socket and
completes the connection.
In \emph{connector mating}, the robot holds one connector
in each hand, aligns them, and brings them into full engagement.
Manipulating both parts requires maintaining their relative
pose throughout the interaction.

Both tasks require continued control input after alignment:
friction resists further motion, and completing the connection
requires sufficient force to engage a retaining latch.
They therefore test whether a policy can sustain the control
demand needed to complete an interaction despite mechanical
resistance. For both tasks, \emph{alignment} requires partial
engagement with a relative pose that permits further motion.
\emph{Full insertion} requires complete seating and engagement
of the latch; partial engagement does not count as final success.

In \emph{package transfer}, the left hand picks up a deformable
package and passes it to the right hand, which places it in
a target tray. \emph{Pickup} requires lifting and holding
the package clear of its support surface. \emph{Placement}
requires releasing it and leaving it within the target tray.
Here, deformability and grasp stability are the main challenges,
providing a complementary setting for examining command
supervision across different interaction demands.

Table~\ref{tab:main-results} reports stage success rates using
all trials for each task as the denominator. Full insertion
and placement rates are therefore not conditioned on completion
of the earlier stage. Stage labels are assessed from complete
evaluation videos independently of CSDW scores.

\begin{figure*}[t]
\centering
\includegraphics[width=\textwidth]{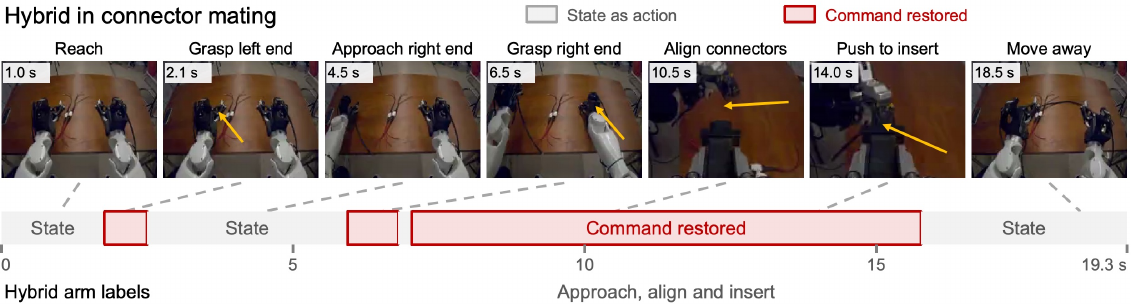}
\caption{Selective command retention in a connector mating demonstration.
Keyframes are linked to the Hybrid label timeline: red intervals use
Command arm targets and gray intervals use State targets. Retained
intervals include brief grasping events and a longer period spanning
approach, alignment, and insertion. Selection uses the 60th percentile of
scores over the training dataset; Command targets occupy 53.5\% of this
demonstration's frames.}
\label{fig:experiment-retention}
\vspace{-1 em}
\end{figure*}

\begin{figure}[t]
\centering
\includegraphics[width=\columnwidth]{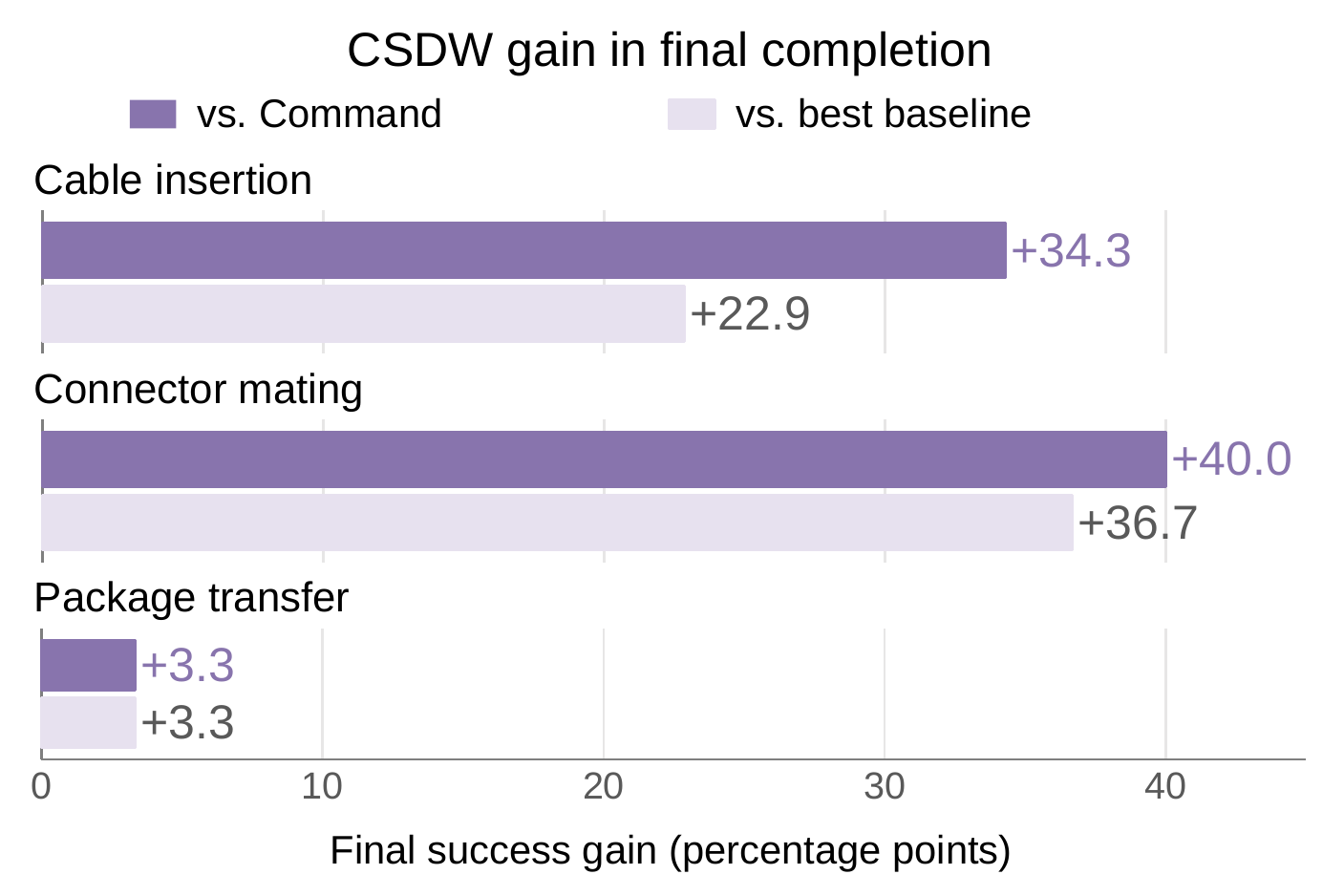}
\caption{CSDW gains in final stage success over Command and the best
baseline, measured in percentage points. The best baseline is Hybrid
for both insertion tasks and State/Command (tied) for package transfer.}
\label{fig:experiment-weighting}
\end{figure}

\subsection{Task and Phase Dependence}
\label{subsec:supervision-comparison}

\emph{Dependence on interaction conditions.}
Fig.~\ref{fig:task_phrase_dependence} summarizes the
Command--State success-rate differences reported in
Table~\ref{tab:main-results}.
Command improves final completion over State by 34.3 percentage
points on cable insertion and 33.3 points on connector mating.
On package transfer, both achieve 80.0\% placement success.
The benefit of retaining commands is therefore larger in the
evaluated tasks involving sustained interaction constraints,
while measured-state supervision remains competitive for
deformable object transfer.

\emph{Variation within a task.}
The supervision gap also varies across assessed stages.
State and Command achieve identical cable alignment success,
while the connector alignment gap is 10.0 percentage points.
Both gaps are smaller than those at final completion
(Fig.~\ref{fig:task_phrase_dependence}(a,b)).
These results show that the advantage of command supervision is larger at full insertion than at alignment.
Package transfer shows a different pattern: State exceeds
Command by 10.0 points at pickup, but this advantage does
not carry through to final placement
(Fig.~\ref{fig:task_phrase_dependence}(c)).

Fig.~\ref{fig:experiment-behavior} illustrates representative
outcomes: State leaves the plug or connectors partially
engaged, while Command achieves full engagement.
After alignment, completing these interactions requires
overcoming frictional resistance and engaging a final latch.
Measured displacement can remain small during this process,
even while the operator continues requesting motion.
Recorded commands retain this continuing control demand,
which may be weakly reflected in measured-state targets.
This provides a plausible explanation for the larger
supervision gaps at completion, although the stage outcomes
alone do not establish the underlying mechanism.

Together, the cross-task and within-task comparisons support
H1: the usefulness of command supervision depends on
interaction conditions and phase, with larger observed
benefits when completing sustained constrained interactions.

\subsection{Local Value of Command Information}
\label{subsec:hybrid-results}

\emph{Preserving the benefit of command supervision.}
Hybrid reaches 71.4\% full insertion success on cable insertion, compared
with 25.7\% for State and 60.0\% for Command. On connector mating,
Hybrid achieves 36.7\%, compared with 0.0\% for State and 33.3\% for
Command. In both tasks, selective retention preserves the observed
performance benefit of full Command supervision over State, while the
remaining arm targets come from measured states. These results support H2: in the two evaluated insertion tasks, retaining commands only in the selected segments preserves the observed advantage over State supervision.

Fig.~\ref{fig:experiment-retention} makes the label selection explicit
for a connector mating demonstration. Retained Command intervals include
brief grasping events and a longer period spanning approach, alignment,
and insertion; other intervals use State targets. The selected region
therefore includes a maintained interaction as well as shorter transitions.
The timeline is obtained from discrepancy scores without task phase
annotations. It illustrates the selected supervision, without establishing
that commands outside these intervals are uninformative.

Hybrid's higher observed final success than Command may result from using State targets during free motion and retaining Command targets during constrained interaction. However, Hybrid's alignment rate is
lower than Command's on cable insertion and unchanged on connector mating.
Its higher final success reflects more consistent completion after
alignment, without establishing smoother motion or better alignment.
On package transfer, Hybrid achieves 76.7\% placement success, close to
the 80.0\% of State and Command, with no corresponding benefit from
selective retention in this task.

\subsection{Temporal Discrepancy Weighting}
\label{subsec:csdw-results}

\emph{Weighting adds value beyond retaining commands.}
Having examined which supervision is retained, we next evaluate how it
is weighted. CSDW keeps all Command targets and uses temporal discrepancy
evidence to determine their additional training emphasis.
CSDW reaches 94.3\% full insertion success on cable insertion and 73.3\%
on connector mating, improving over Command by 34.3 and 40.0 percentage
points, respectively. Since CSDW and Command retain the same targets,
these gains support H3: temporal discrepancy can be used to improve
training after the command information is already present in the targets.

Fig.~\ref{fig:experiment-weighting} summarizes these gains relative to
both Command and the best baseline for each task. The insertion gains
remain substantial even relative to Hybrid, the strongest baseline on
these tasks. By comparison, package placement reaches 83.3\%, a gain of
3.3 percentage points over Command and the best baseline. The larger
observed gains on the insertion tasks are consistent with the task
dependence of command information observed in the preceding comparisons.

\emph{Improvements extend to alignment.}
CSDW also raises alignment success over Command by 17.1 percentage
points on cable insertion and 36.7 points on connector mating.
Alignment and full insertion rates coincide for CSDW in both tasks,
indicating gains in reaching an insertion-ready configuration as well
as completing engagement. The benefit therefore extends beyond continuing
an insertion request after alignment.

This pattern is consistent with the temporal structure of CSDW weights.
Alongside delayed progress and sustained effort, the effort changes profile
emphasizes increases and decreases in corrected effort, which may accompany
the onset or relaxation of a maintained request. Additional supervision
around such transitions
may help the policy learn when to adjust its pose, maintain an insertion
request, or reduce it, complementing supervision throughout the interaction.
The alignment gains are measured for the complete method and do not
isolate the contribution of effort changes. Likewise, because the weights
are not normalized to unit mean, the comparison with Command evaluates
their combined effect on supervision allocation and overall loss scale.

\vspace{ 1 em}
\section{Conclusion}

Our findings show that command information provides benefits beyond
state-as-action supervision that depend on interaction conditions
and phase. These benefits are more pronounced under sustained
constraints, while state-as-action remains competitive on package
transfer. Preserving the observed benefit through selective command
retention supports the local value of this information. CSDW further
demonstrates its practical use: temporal command--state discrepancy
can guide training weights that improve completion over uniform
command supervision on constrained tasks, without changing policy
architecture or inference.

More broadly, our findings highlight the importance of keeping
command information available throughout robot data pipelines.
When action targets are reconstructed from measured motion or
keyframe poses~\cite{walke2023bridgedata,octo2024,shridhar2022peract},
the resulting labels may no longer retain this information.
We therefore encourage preserving recorded commands and measured
states alongside processed training labels, with clear documentation
of their temporal alignment, control interface, and label transformations.
To further study when this distinction matters, simulation benchmarks
should expose both command and response streams and evaluate
supervision choices under varied controller dynamics and contact
constraints.

\vspace{-0.5 em}

\bibliographystyle{IEEEtran}
\bibliography{reference}

\end{document}